\documentclass[sigconf]{acmart}

\AtBeginDocument{%
  \providecommand\BibTeX{{%
    \normalfont B\kern-0.5em{\scshape i\kern-0.25em b}\kern-0.8em\TeX}}}

\usepackage{graphicx}
\usepackage{subfigure}
\usepackage{xspace}
\usepackage{multirow}
\usepackage{hyperref}
\usepackage[ruled,vlined,linesnumbered]{algorithm2e}

\newtheorem{defn}{Definition}

\newcommand{\eat}[1]{}
\ifodd 1

\newcommand{\TODO}[1]{{\color{red}TODO:{#1}}}

\else

\newcommand{\TODO}[1]{}

\fi

\newcommand{\kw}[1]{{\ensuremath {\mathsf{#1}}}}
\newcommand{\polestar}{\kw{Polestar}\xspace}

\newcommand{\shanghai}{{\sc Shanghai}\xspace}
\newcommand{\guangzhou}{{\sc Guangzhou}\xspace}

\newcommand{\etal}{\emph{et~al.}\xspace}
\newcommand{\eg}{\emph{e.g.},\xspace}
\newcommand{\ie}{\emph{i.e.},\xspace}

\newcommand{\aka}{\emph{a.k.a.}\xspace}
\newcommand{\etc}{\emph{etc.}\xspace}

\newcommand\figref[1]{Figure~\ref{#1}}
\newcommand\algoref[1]{Algorithm~\ref{#1}}
\newcommand\tabref[1]{Table~\ref{#1}}

\eat{
\copyrightyear{2020} 
\acmYear{2020} 
\setcopyright{acmcopyright}\acmConference[KDD '20]{Proceedings of the 26th ACM SIGKDD Conference on Knowledge Discovery and Data Mining}{August 23--27, 2020}{Virtual Event, CA, USA}
\acmBooktitle{Proceedings of the 26th ACM SIGKDD Conference on Knowledge Discovery and Data Mining (KDD '20), August 23--27, 2020, Virtual Event, CA, USA}
\acmPrice{15.00}
\acmDOI{10.1145/3394486.3403281}
\acmISBN{978-1-4503-7998-4/20/08}
}
\begin{document}
\fancyhead{}

%
\title{Polestar: An Intelligent, Efficient and National-Wide \\
Public Transportation Routing Engine}

%
\author{Hao Liu$^{1}$, Ying Li$^{2}$, Yanjie Fu$^{3}$, Huaibo Mei$^{2}$, Jingbo Zhou$^{1}$, Xu Ma$^{2}$,  Hui Xiong$^{4*}$}\thanks{$^*$Corresponding author.}
\affiliation{
$^1$Business Intelligence Lab, Baidu Research,
$^2$Baidu Maps, Baidu Inc.,
$^3$University of Central Florida,
$^4$Rutgers University\\
$^{1,2}$\{liuhao30, liying, meihuaibo, zhoujingbo, maxu\}@baidu.com,
$^3$yanjie.fu@ucf.edu,
$^4$hxiong@rutgers.edu
}

%
\begin{abstract}
Public transportation plays a critical role in people's daily life.
It has been proven that public transportation is more environmentally sustainable, efficient, and economical than any other forms of travel~\cite{litman2015evaluating,beirao2007understanding}.
However, due to the increasing expansion of transportation networks and more complex travel situations, people are having difficulties in efficiently finding the most preferred route from one place to another through public transportation systems.  
To this end, in this paper, we present \polestar, a data-driven engine for intelligent and efficient public transportation routing.
Specifically, we first propose a novel Public Transportation Graph~(PTG) to model public transportation system in terms of various travel costs, such as time or distance. Then, we introduce a general route search algorithm coupled with an efficient station binding method for efficient route candidate generation.
After that, we propose a two-pass route candidate ranking module to capture user preferences under dynamic travel situations. Finally, 
experiments on two real-world data sets demonstrate the advantages of \polestar in terms of both efficiency and effectiveness.
Indeed, in early 2019, \polestar has been deployed on Baidu Maps, one of the world's largest map services.
To date, \polestar is servicing over 330 cities, answers over a hundred millions of queries each day, and achieves substantial improvement of user click ratio.
\end{abstract}

%
%
\maketitle

{\fontsize{8pt}{8pt} \selectfont
\textbf{ACM Reference Format:}\\
Hao Liu, Ying Li, Yanjie Fu, Huaibo Mei, Jingbo Zhou, Xu Ma, Hui Xiong. 2020. Polestar: An Intelligent, Efficient and National-Wide Public Transportation Routing Engine. In \textit{the 26th ACM SIGKDD Conference on Knowledge Discovery and Data Mining (KDD’20), August 23--27, 2020, Virtual Event, USA.} ACM, New York, NY, USA, 9 pages. 
https://doi.org/10.1145/3394486.3403281}

\section{Introduction}
Public transportation is a form of transit that offers people travel together along designated routes.
In recent decades, public transportation has become ubiquitous, and we have witnessed the rapid increase in access to and use of the public transportation system. 
For example, in 2018, the China government invested more than $\$74$ billion\footnote{http://www.xinhuanet.com/english/2019-02/13/c\_137819050.htm} in public transportation infrastructure and the public transportation systems took over $120$ billion trips\footnote{http://xxgk.mot.gov.cn/jigou/zhghs/201905/t20190513\_3198918.html}.
In fact, public transportation is playing a key role in the daily life of urban residents.
Public transport modes such as bus, metro, and light rail can help reduce urban traffic jam, improve urban transportation network efficiency, and ultimately reduce urban commute costs~\cite{litman2015evaluating,beirao2007understanding}.

However, despite the popularity and various advantages of public transportation, it is challenging for users to find the most preferred routes from a variety of routes, because of the complex public transportation networks, new emerging public transportation tools~(\eg vanpooling, on-demand ride-hailing, shared bike, \etc), and the dynamic travel context~(\eg transportation station distribution, weather, travel intention, \etc).
As a result, public transportation routing services such as Baidu Maps and Google Maps become essential tools in people's daily lives.

\begin{figure}[t]
    \begin{minipage}{1.\linewidth}
        \centering
        \subfigure[{\small Ranked route list.}]{\label{fig:product}
        \includegraphics[width=0.4\textwidth]{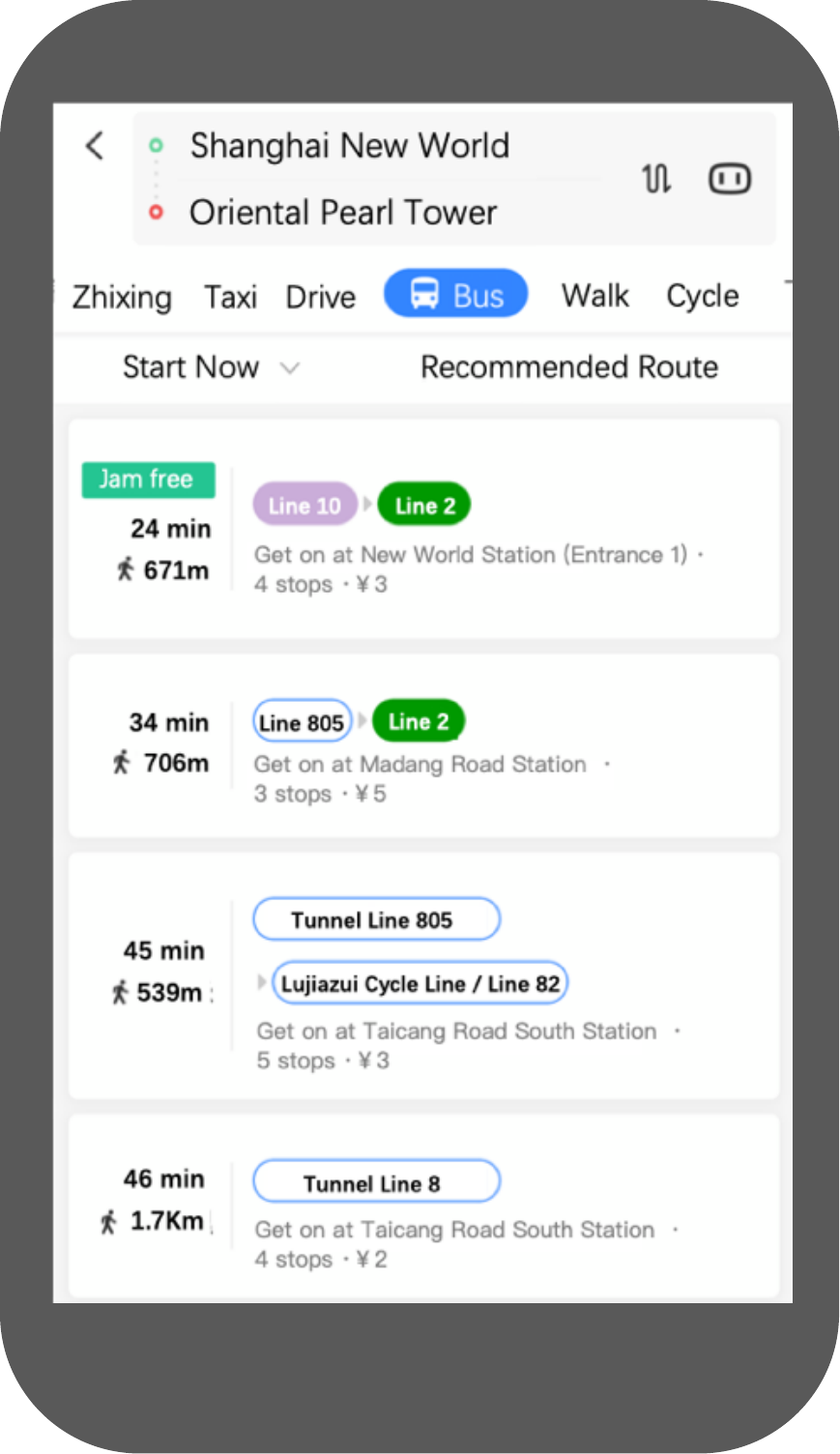}}
        \hspace{5pt}
        \subfigure[{\small Details of the first route.}]{\label{fig:route}
        \includegraphics[width=0.4\textwidth]{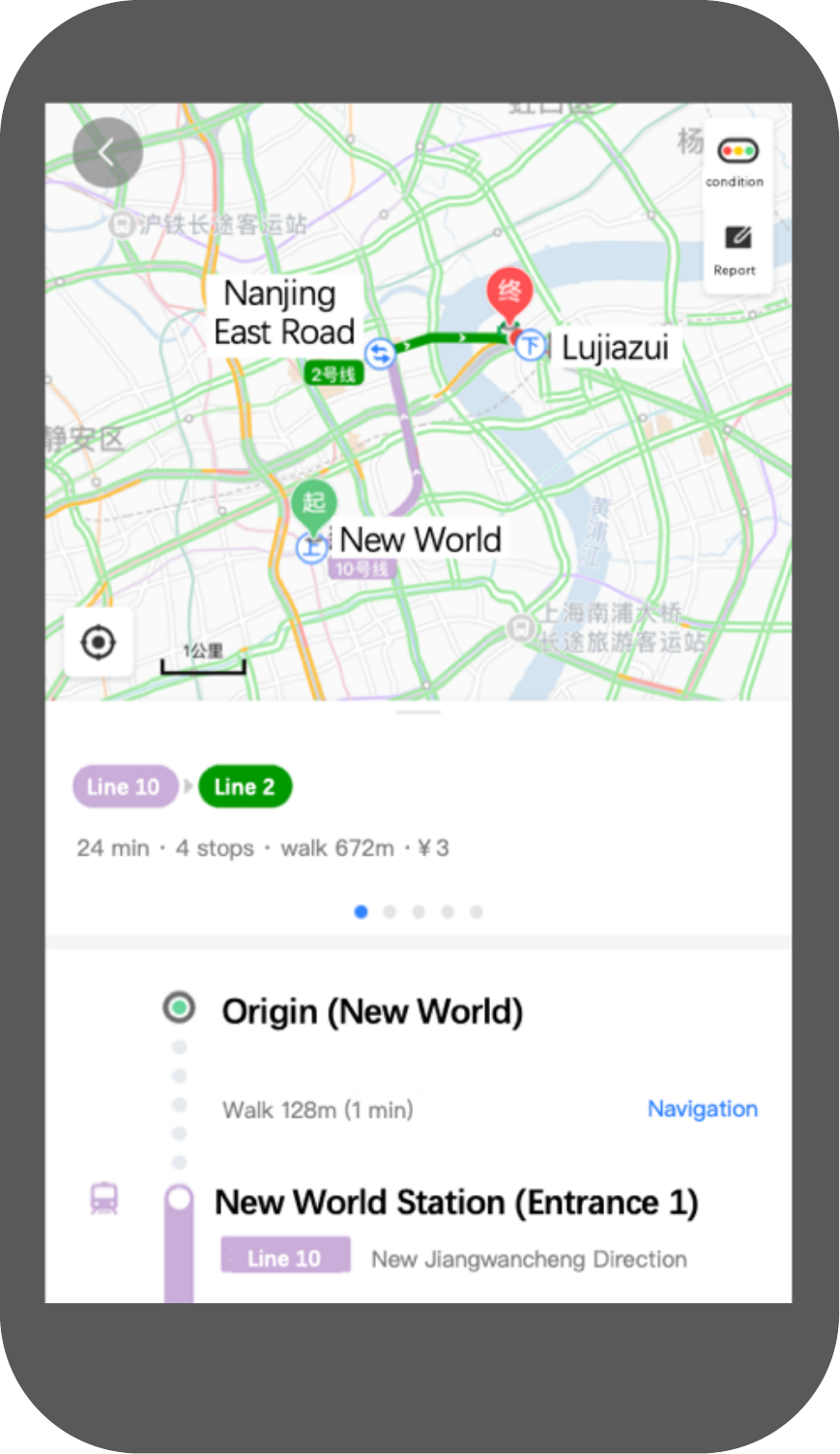}}   
    \end{minipage}
    \vspace{-15pt}
    \caption{
    Examples of user interfaces of \polestar. 
    (a) A ranked list of public transportation routes from Shanghai New World to the Oriental Pearl Tower, two landmarks in Shanghai, China. (b) The details of the first route in (a), which is a metro based transportation route. The first route is fastest and is with least number of transfer.}
    \vspace{-10pt}
    \label{fig:product}
\end{figure}

While geographic routing is well-studied, the predominant research and applications are mostly about routing with road networks, only a few works focus on public transportation routing.
For example, Efentakis \etal~\cite{efentakis2016scalable} formulated the public transportation routing problem as a database query and proposed a pure SQL based routing framework.
Wang \etal~\cite{wang2015efficient} and Delling \etal~\cite{delling2015public} modeled public transportation networks as a timetable graph and proposed a labeling based index to speedup shortest path queries. 
However, all the above approaches mainly focus on city-wide public transportation routing, and only consider single or a few transport modes~(bus and metro).
More importantly, all the above approaches focus on optimizing static criteria~(\eg earliest arrival, latest departure, shortest travel time), but overlook the user preference under dynamic situational context, which is important for user decision making.
For example, metro may be a more preferable choice during morning rush hour or under severe weather condition, whereas the cheapest route may be a better choice when one's trip purpose is not in an emergency.

In fact, building a public transportation routing engine has far beyond searching shortest paths. The major challenge comes from two aspects.
First, the rapid expansion of the public transportation network induces highly overlapped transportation lines. The alternative sub-routes and the ultimate line transfer lead to the combinatorial explosion of the route search space.
The first challenge is how we can efficiently generate feasible route candidates.
Second, the user preference is highly dynamic and depends on many factors such as price, time period, and weather condition. Simply sorting routes based on static criteria such as distance or time fails to deliver a satisfactory user experience. So, the second challenge is how to rank route candidates by characterizing user preference under dynamic travel context.

To tackle above challenges, in this paper, we present \polestar, an intelligent public transportation routing engine. We hope to share our practical experience on how to build an intelligent, efficient, and national-wide public transportation routing service.
In early 2019, \polestar has been deployed on Baidu Maps, one of the world's largest navigation apps, servicing over 330 cities in mainland China.
\figref{fig:product} shows the user interface of \polestar on Baidu Maps app.

\textbf{Contributions}.
To the best of our knowledge, this is the first work introducing a deployed national-wide public transportation routing engine that answers a hundred millions of queries each day.
Specifically, we first propose the public transportation graph, PTG, that assembles heterogeneous public transportation lines into a unified structure. PTG elegantly models various travel costs and reduces routing complexity by mapping public transportation lines into a set of physical and virtual graphs.
Beside, we design efficient route candidate generation algorithms, coupled with an efficient station binding method. In average, route candidate generation can be done in tens of milliseconds in \polestar.
Moreover, we propose a two-pass route candidate ranking pipeline to capture user preference under dynamic travel context. The route candidate ranking pipeline achieves $9.4\%$ relative improvement of user click ratio in the production environment.
Finally, we develop a series of optimization techniques to reduce web service latency and discuss several deployment issues. Extensive experiments on urban-scale real-world datasets show that Polestar achieves less than $250$ms latency in average and exhibits excellent ranking performance compared with six baselines.

\section{Framework}\label{sec:framework}

\begin{figure}[t]
    \centering
    \hspace{-3ex}
    \includegraphics[width=.46\textwidth]{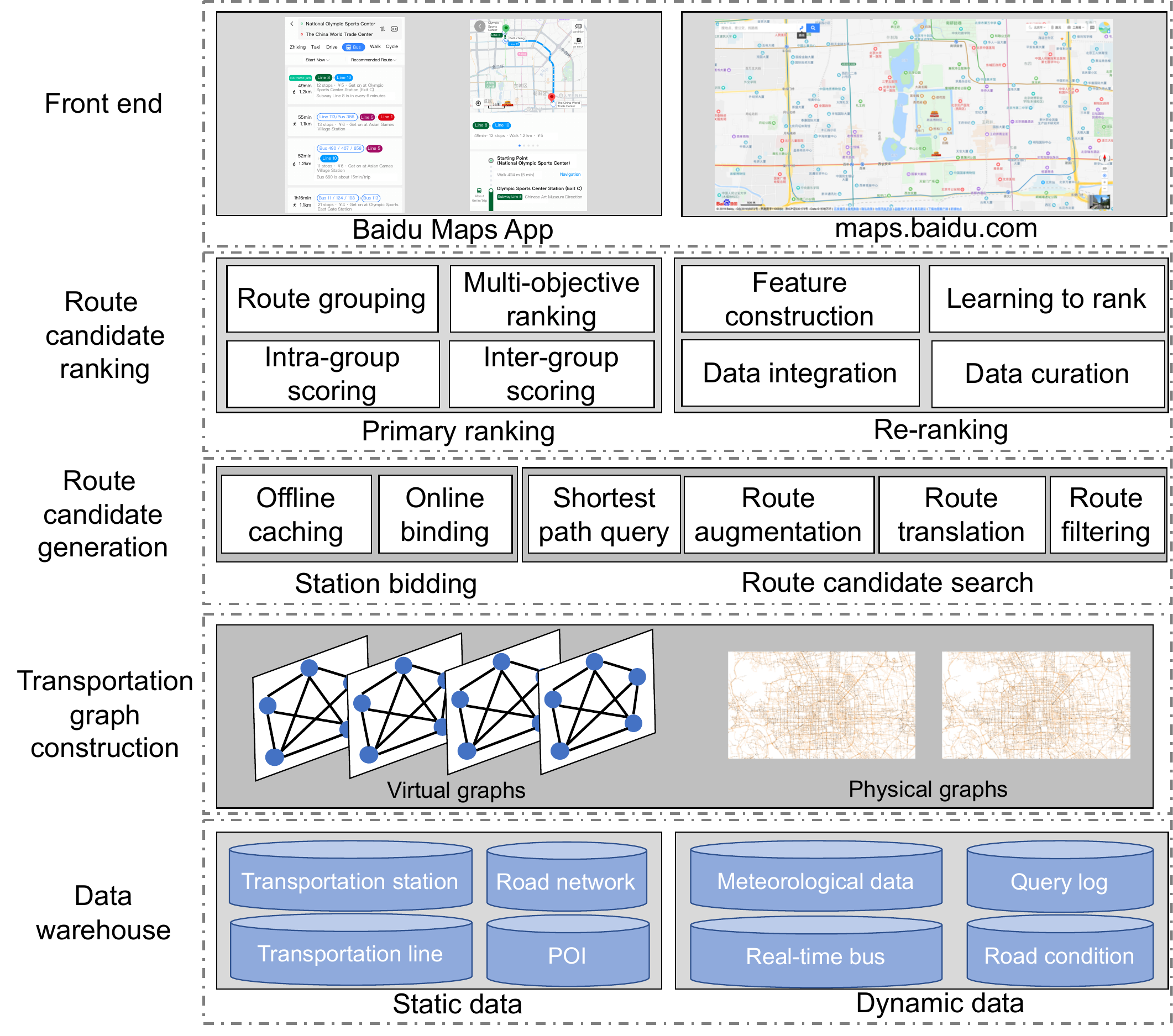}
    \vspace{-5pt}
    \caption{The framework overview of \polestar.}
    \vspace{-15pt}
    \label{fig:framework}
\end{figure}

\figref{fig:framework} shows the framework of \polestar, which consists of five components, the \emph{Data warehouse}, the \emph{Public transportation graph construction}, the
\emph{Route candidate generation}, the \emph{Route candidate ranking} and the \emph{Front end}.
The \emph{Data warehouse} stores a wide range of datasets on a distributed cluster, such as the transportation station data and the transportation line data. 
Based on transportation line related datasets in the \emph{Data warehouse}, the \emph{Graph construction} module compiles the PTG, which including a set of intra-city graphs.
When a routing query is submitted, the \emph{Route candidate generation} module searches a set of feasible route candidates based on the PTG.
Concretely, we first bind the origin and the destination to proper stations based on a pre-computed station cache. After that, a bidirectional shortest path search algorithm is applied on a set of virtual graphs simultaneously to generate set of feasible route candidates, which later will be translated to human-readable routes based on the physical graph.
Once a set of route candidates is obtained, a two-pass \emph{Route candidate ranking} module is invoked for context-aware ranking.
Specifically, the primary ranking first partitions route candidates into several route groups and then select a small subset of diversified route candidates.
After that, the re-ranking step constructs a rich set of features and applies a machine learning based model to decide the final rank of each route candidate. 
Finally, the ranked route list is returned to the \emph{Front end}.
There are two interfaces in the \emph{Front end}: the App interface for mobile devices and the webpage interface for PC.

\begin{figure*}[t]
    \begin{minipage}{1.0\linewidth}
        \centering
        \subfigure[{\small Spatial distribution of destinations}]{\label{fig:query-destination}
        \includegraphics[width=0.24\textwidth]{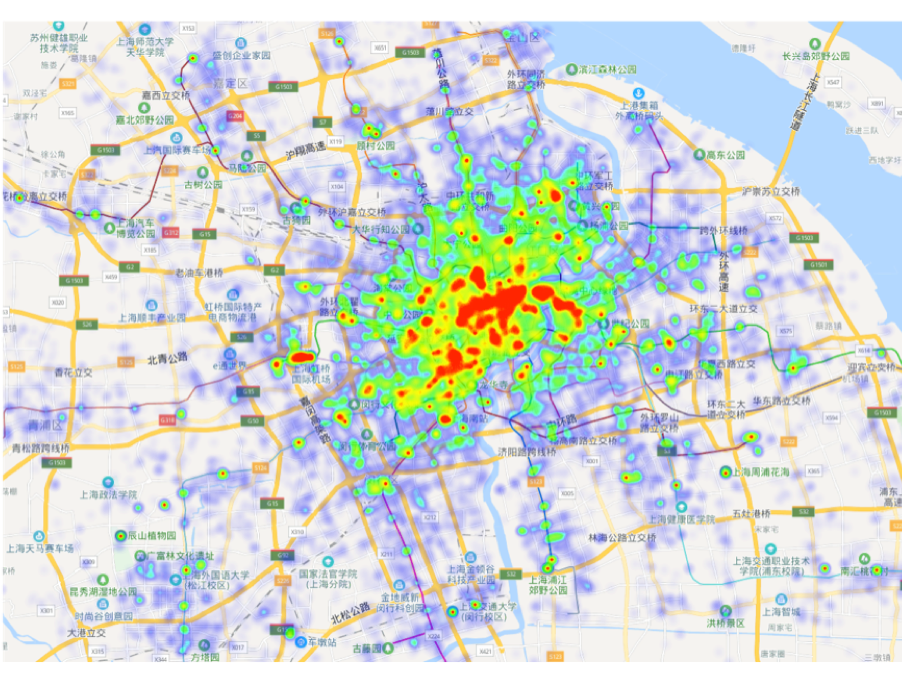}}
        \subfigure[{\small Feedback distribution}]{\label{fig:click-distribution}
        \includegraphics[width=0.24\textwidth]{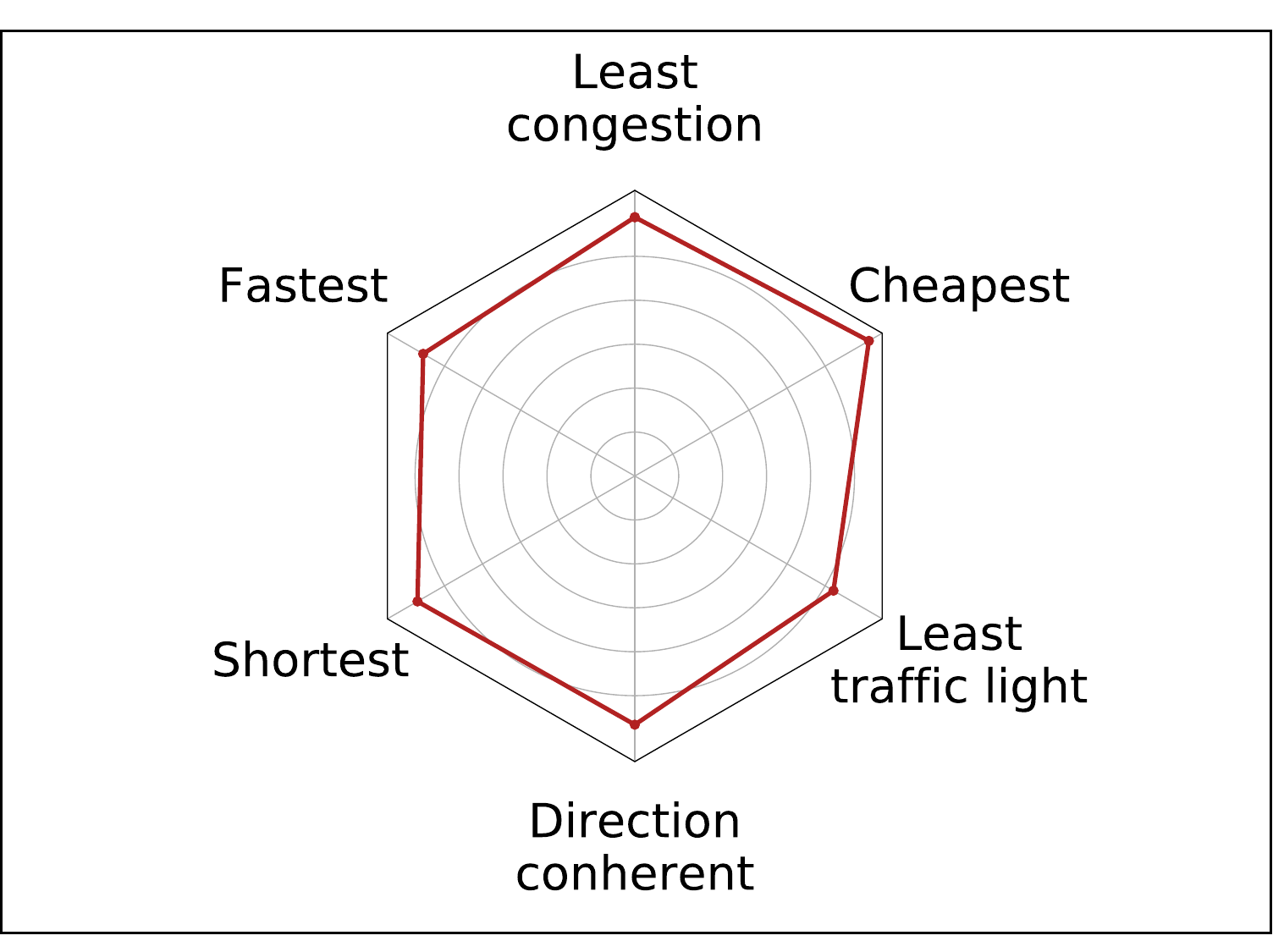}}
        \subfigure[{\small Distance distribution}]{\label{fig:distance-feedback}
        \includegraphics[width=0.24\textwidth]{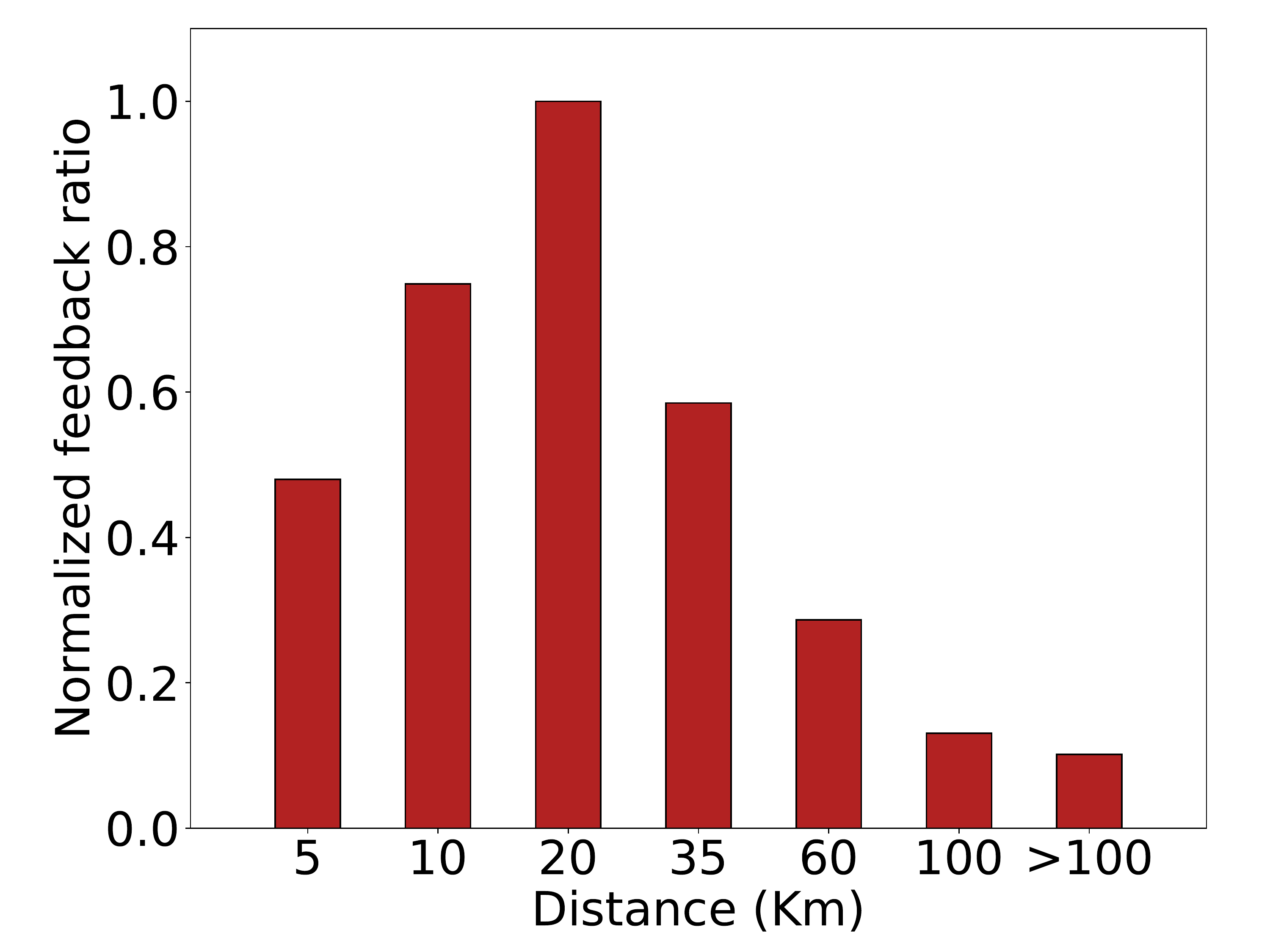}}
        \subfigure[{\small Day distribution}]{\label{fig:hour-feedback}
        \includegraphics[width=0.24\textwidth]{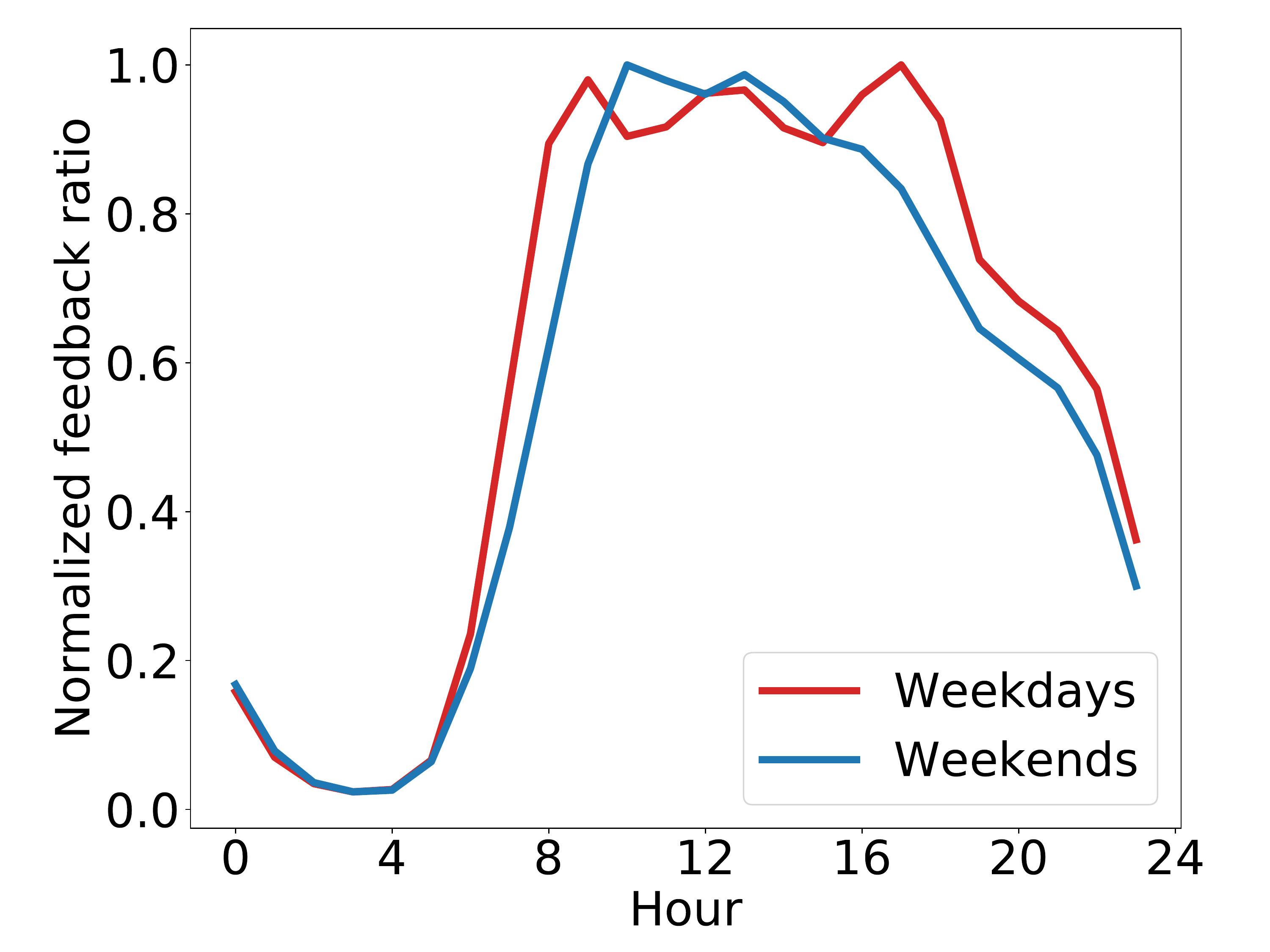}}
        \vspace{-3ex}
    \end{minipage}
    \caption{Distributions of the \shanghai dataset: (a) the spatial distribution of query destinations; (b) the distribution of routes with user feedbacks; (c)the distribution of travel distances; (d) the distribution of travel time~(hour).}
    \vspace{-2ex}
    \label{figs:data-distribution}
\end{figure*}

\section{Data description and analysis}\label{sec:data}
This section introduces datasets used in \polestar, with a preliminary data analysis. In this paper, we use two datasets, \shanghai and \guangzhou. 
Both of them are randomly sampled from 60 consecutive days in early 2019. The statistics of two datasets are summarized in Table~\ref{table-data}.
\begin{table}[t]
\centering
\begin{small}
\caption{Statistics of datasets.}
\vspace{-3ex}
\begin{tabular}{c c | c  c} \hline
\multicolumn{2}{c|}{\bf{Data description}} & \shanghai & \guangzhou  \\ \hline \hline
\multirow{1}{*}{Query log data} &\# of sessions & 5,900,463 &2,750,316  \\  \hline
\multirow{4}{*}{Geographical data}  & \# of stations & 68,687  & 65,525 \\
  & \# of lines & 3,341  & 2,839 \\ 
  & \# of road segments &  406,195 & 284,168 \\ 
  & \# of POIs & 1,594,684 & 982,059 \\ \hline
Meteorological data & \# of records & 23,424  & 16,104  \\ \hline
\end{tabular}
\label{table-data}
\vspace{-5ex}
\end{small}
\end{table}

\textbf{Geographical data}. 
We use large-scale geographical datasets to build \polestar, including: (1) the transportation station data, (2) the transportation line data, (3) the road network data, and (4) the point of interest~(POI) data~\cite{zhou2019collaborative}.
All geographical data are collected from (i) professional surveyors employed
by Baidu Maps, (ii) the crowdsourcing platform in Baidu.
Public transportation stations are fundamental data to build the PTG.
Besides, geographical data contains rich semantic information regarding user mobility~\cite{zhang2020semisupervised,zhou2013semi}. For example, regional transportation station density influences the public transportation accessibility of a specific area, and the POI type of destination reflects user travel intention~\cite{wang2017kdd}.

\textbf{Query log data}. Query log data captures user interactions with Baidu Maps.
According to the user interaction loop, the query log data generated during user-App interactions can be further categorized into \textit{query records}, \textit{routes records}, and \textit{feedback records}. 
Briefly, a query record indicates a route query from a user on Baidu Maps, a routes record contains a list of feasible candidate routes present to the user, and a feedback record represents user's preference of given candidate routes.
Different from traditional recommenders~\cite{graepel2010web} collect user clicks as feedback, we extract more reliable feedbacks to distinguish better routes, which include \emph{add route to favorites}, \emph{share route with others}, \emph{screenshot} and \emph{navigation}.

\textbf{Meteorological data}. Meteorological conditions are critical factors for trip planning. For example, a route with less walk distance is preferable in the case of snow, rain, and severe air pollution. 
Each meteorological data point consists of a location, a timestamp, the weather, the temperature, the wind strength, the wind direction, and the Air Quality Index (AQI).
We use the meteorological data from an online meteorology website of the Chinese government.

\textbf{Data analysis}. To further understand the distributions of each dataset, we use the dataset \shanghai for illustration. 
We observe similar data distributions in other cities and time periods.
\figref{fig:query-destination} shows the distributions of destinations in query records. 
As can be seen, most destinations fall in the downtown areas in \shanghai.
\figref{fig:click-distribution} plots the distribution of user-preferred routes~(\ie route with user feedbacks), the ratio of least congestion, fastest, shortest, direction coherent, least traffic light and cheapest are $58.9\%$, $55.6\%$, $57.1\%$, $56.6\%$, $52.2\%$ and $61.5\%$, respectively. Since a route may be the best in multiple aspects~(\eg least congestion and fastest), the overall ratio is greater than $100\%$. Overall, we observe multiple factors users may concern when planning a trip.
\figref{fig:distance-feedback} depicts the spatial distribution of user feedbacks. We observe over $80\%$ trips are within $35$Km and trips of the distance around $20$Km are most popular, which provides extra information for public transportation routing. 
Finally, \figref{fig:hour-feedback} shows the temporal distribution of user feedbacks. Overall, we observe the user feedback ratio at daytime is higher than night, and the distribution on workday and weekend are also different.
Above observations motivate us to incorporate multi-source urban data to model the dynamic travel context, and build a machine learning based model for intelligent route recommendation.

\section{Public transportation graph construction}\label{sec:graph-construction}
One major design objective of \polestar is to provide an efficient national-wide public routing service.
However, it is computationally expensive to route on a unified public transportation graph~\cite{efentakis2016scalable,wang2015efficient} that contains millions of transportation lines.
In this section, we propose the \emph{public transportation graph} to reduce the route candidate generation complexity on various travel costs.
Specifically, the unified graph structure partitions transportation lines into a set of disjoint intra-city publication transportation graphs, the route candidate search space is therefore bounded into a relatively smaller range.
Moreover, the public transportation graph decouples each subgraph into one \emph{physical graph} and multiple \emph{virtual graphs}, where each virtual graph corresponds to a different travel cost.

\begin{figure}[t]
    \begin{minipage}{1.0\linewidth}
        \centering
        \subfigure[{\small Transportation lines on the map.}]{\label{fig:raw_graph}
        \includegraphics[width=0.48\textwidth]{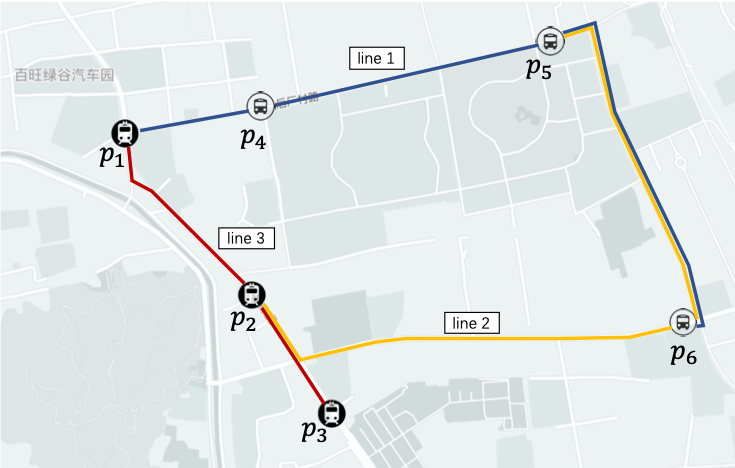}}
        \subfigure[{\small Physical transportation graph.}]{\label{fig:physical_graph}
        \includegraphics[width=0.48\textwidth]{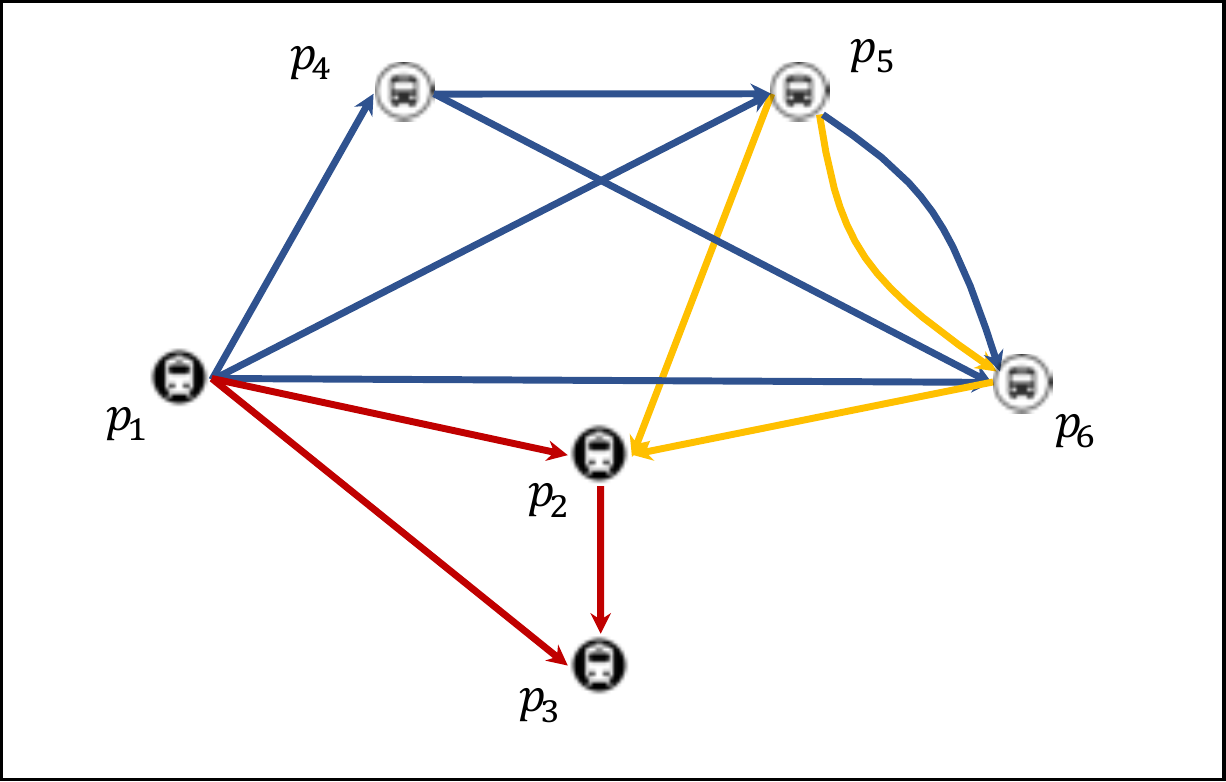}}
        \vspace{-3ex}
    \end{minipage}
    \caption{
    Example of transportation graph construction.
    }
    \vspace{-5ex}
    \label{fig:graph_construction}
\end{figure}

Consider a set of public transport modes $M=\{m_1, m_2, \dots, m_i\}$.
A \emph{physical transportation station} $p_i\in P$ is represented as a geographical coordinate $(\varphi_i, \lambda_i)$. Note that $p_i$ can be passed by multiple transportation lines.
A  \emph{Transportation line} is defined as a tuple $(m_i, l_i)$, where $m_i \in M$ is a public transport mode,  $l_i=p_1 \rightarrow p_2 \rightarrow \cdots \rightarrow p_n$ is an ordered physical transportation station list.
In practice, a transportation line may be a bus line, a metro line, a ferry route, etc.

\begin{defn}(\textbf{Physical transportation graph})
A physical transportation graph is a 5-tuple $G^P=(P, E, O, D, L_{E})$, where $P$ is a set of physical transportation stations, $E$ is a set of edges between physical transportation stations, $O$ is a mapping set $P\rightarrow E$ assigning to edge its origin station, $D$ is a mapping set $P\rightarrow E$ assigning to edge its destination station, and $L_{E}$ is a mapping set marks the transportation line of each edge.
\end{defn}

The physical transportation graph is a directed multi-graph, where each edge is with its own identity.
Give two physical transportation stations $p_i$ and $p_j$, there is an edge from $p_i$ to $p_j$ if and only if there exists a transportation line pass from $p_i$ to $p_j$ without transfer. Note that there may have multiple edges from $p_i$ to $p_j$ if there are multiple transportation lines pass from $p_i$ to $p_j$.
Consider a set of transportation lines shown in \figref{fig:raw_graph}, \figref{fig:physical_graph} depicts the corresponding physical transportation graph. For example, as line 3 passes $p_1$, $p_2$ and $p_3$, there are three edges $(p_1,p_2)$, $(p_1,p_3)$ and $(p_2,p_3)$ labeled as line 3. Besides, there are two transportation lines that pass from $p_5$ to $p_6$, from which we derive two edges between the corresponding nodes.

The advantages of the physical transportation graph are two-fold. First, it fits heterogeneous transportation lines into a unified graph representation, which eases subsequent route candidate generation. Second, directly connecting transportation stations in the same transportation line significantly reduces the search depth in route generation, therefore increasing the number of feasible route candidates in fixed search depth.

Similar to the physical transportation station, we define \emph{virtual transportation station} $v_i\in V$ based on the physical transportation station, each of which corresponds to an individual transportation line. 
In other words, a physical transportation station is mapped to multiple virtual transportation stations, and each virtual transportation station corresponds to a different transportation line.

\begin{defn}(\textbf{Virtual transportation graph})
A virtual transportation graph is defined as a 4-tuple $G^V=(V,E, S_{PV}, c_E)$, where $V$ is a set of virtual transportation stations, $E$ is a set of edges between virtual transportation stations, $S_{VP}$ is a table $V\rightarrow P$ which maps virtual transportation stations to physical transportation stations, and $c_E$ is a mapping set describes the weight of each edge.
\end{defn}

The relation between virtual transportation station and physical transportation is many-to-one.
For example, given the physical transportation graph shown in \figref{fig:physical_graph}, virtual station $v^{p_5}_1$ and $v^{p_5}_2$ are mapped to physical transportation station $p_5$. Two edges between $p_5$ and $p_6$ are mapped to two disjoint virtual edges $(v^{p_5}_1,v^{p_6}_1)$ and $(v^{p_5}_2,v^{p_6}_2)$, associated with different travel cost.
In \polestar, a virtual transportation graph is stored in two parts, the station mapping table, and the virtual graph table. 

Currently, each physical transportation graph corresponds to three virtual transportation graphs, \ie the distance graph, the travel time graph, and the walk distance graph. 
All three virtual graphs are isomorphism, except that the weight of each edge is computed from different cost function.
The weight of each edge is estimated from query log data and historical road conditions.
Note that other types of virtual graphs can be extended on demand.

Based on the physical graph and virtual graphs for public transportation networks in each city, we construct the \emph{public transportation graph}.

\begin{defn}(\textbf{Public transportation graph (PTG)})
A PTG is defined as a set of physical transportation graphs and virtual transportation graphs $G^H=\{G^P_1,G^P_2,\dots,G^V_1,G^V_2,\dots\}$. PTG partitions the public transportation graph into a set of disjoint public transportation graphs. Each sub-graph corresponds to an intra-city public transportation network.
\end{defn}

PTG partitions large-scale graphs into a set of subgraphs and therefore reduces the candidate route generation complexity.

\section{Route candidate generation}\label{sec:routing}
In this section, we describe the detailed process of route generation, including station binding and route candidate search.

\subsection{Station binding}

\begin{figure}[t]
		\centering
        \includegraphics[width=.36\textwidth]{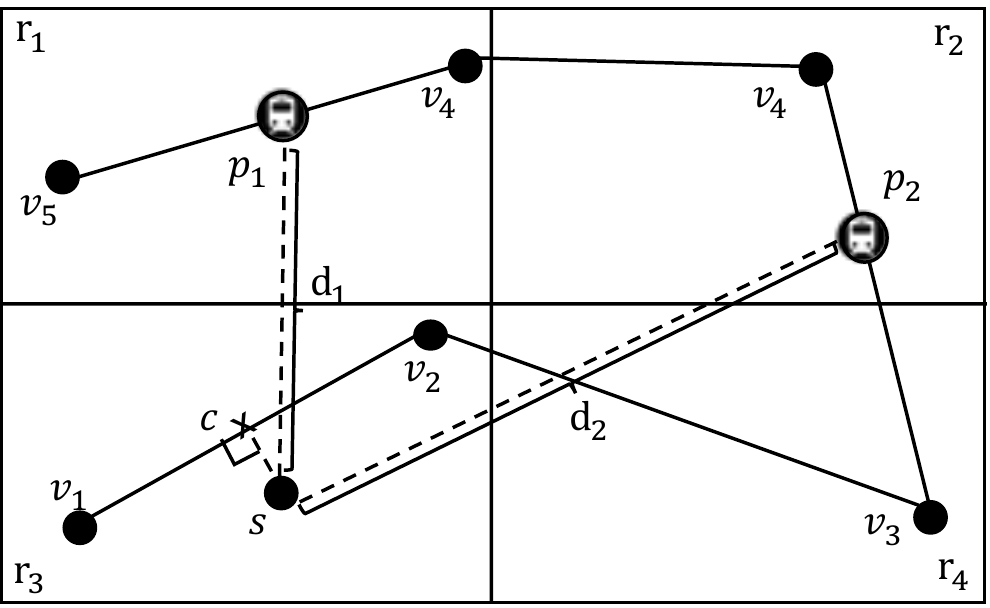}
        \vspace{-3pt}
        \caption{Example of station binding. $s$ is the current location, $u_1$-$u_6$ are road intersections, $p_1$-$p_2$ are physical transportation stations, $r_1$-$r_6$ are partitioned grids, $c$ is projected point from $s$ to $(v_1,v_2)$.}
        \label{fig:station_binding}
        \vspace{-5pt}
\end{figure}

In general, the origin and the destination of a query are arbitrary geographical locations.
A primary step for route candidate generation is binding the origin~(resp. destination) location to physical transportation stations the user can get on~(resp. get off).
A straightforward approach is computing the Euclidean distance between the origin (resp. destination) location with each surrounding physical station and select top-$k$ nearest stations as getting on~(resp. getting off) stations.
However, as the distance between the location and each station is constrained by the pedestrian road network, the actual road network distance may be much longer than the point-to-point Euclidean distance, which may lead to a sub-optimal binding result. 
For example, in \figref{fig:station_binding}, the Euclidean distance from location $s$ to $p_1$ is less than from $s$ to $p_2$, however, the road network distance from location $s$ to $p_1$ is greater than from $s$ to $p_2$.
Another option would be to map the location to a road segment and then compute the road network distance on the fly.
However, this approach leads to great online computation overhead and induce efficiency degradation.
In \polestar, we employ a caching method to bind geographical locations to stations more accurate and efficient. 

The proposed station binding method builds an online-offline framework as follow.
Given a location $s$, a road network $G^R=(V,E)$, and stations $p_i \in P$, we first build an offline nearest station cache.
Specifically, we partition the city into a set of disjoint grid $r_k\in R$, and place each road intersection $u_i\in V$ on this grid.
For each road intersection $u_i$, we apply Dijkstra's algorithm to select a set of stations such that $d(u_i,p_j)<\lambda$, where $d(\cdot)$ is the road network distance and $\lambda$ is a maximum distance threshold.
Note that we project all stations to the nearest road segment.
We then group the distance information of each station by region and stored them in the cache.
In the online binding process, we first map the location $s$ to the nearest POI.
Each POI is projected to a reachable road segment associated with a walking distance.
Take \figref{fig:station_binding} again for example, $s$ is projected to $(u_1,u_2)$ at $c$, the overall distance from the location $s$ to a station $p_1$ can be derived as
\begin{equation}~\label{equ:distance}
d(s, p_1) = d(s,c) + d(c,u_2) + d(u_2,p_1),
\end{equation}
where $d(s,c)$ is the walk distance from location $s$ to road segment, $d(c,u_2)$ is the road network distance from the projected point $c$ to road intersection $u_2$, and $d(u_2,p_1)$ is the road network distance pre-stored in cache.
As $d(s,p_1)>d(s,p_2)$, $p_2$ is selected as the best match for the station.

\subsection{Route candidate search}\label{sec:route-generation}
Given origin and destination stations, we model the route candidate search as a vertex-to-vertex shortest path search problem. 
The intra-city routing generates route candidates in three steps. First, a bidirectional Dijkstra's algorithm~\cite{mohring2007partitioning} is applied to each virtual graph to generate feasible route candidates. The search procedure stops when a criterion is satisfied, \eg maximum search time or the maximum number of route candidates.
Second, route candidates are mapped to physical transportation stations and physical transportation lines via the route translation module. Detailed route information such as price, ETA, route shape is attached to each route candidate.
Third, a route filter and route augmentation module are invoked to augment route candidate diversity and filter out less competitive route candidates~(\eg routes containing cycles). For route augmentation, a greedy algorithm is applied to replace each route segment. Recall the the physical transportation graph is a multi-graph, route augmentation replaces one edge with same origin and destination station in each step. 
The detailed procedure of route candidate search is shown in \algoref{alg:intra}. 

\begin{algorithm}[t]
    \caption{Route candidate search} \label{alg:intra}
    \KwIn{Physical graph $G^p$, virtual graphs $G^v_1, G^v_2, \dots$, origin station $o$, destination station $d$, maximum search time $t$, maximum route candidate set size $k$.}
    \KwOut{$C$: a set of feasible route candidates.}
    Set $C \gets \emptyset$\;
    \While {$|OpenSet| \leq k$ or $running\_time \leq t$}{
        \For {each virtual graph $G^v_i$}{
            $virtual\_routes$ $\leftarrow$ BidirectionalDijkstra($G^v_i$, $o$, $d$)\;
            $physical\_routes$ $\leftarrow$ Translation($G^p$, $virtual\_routes$)\;
            $candidate\_routes$ $\leftarrow$ Augmentation($G^p$, $physical\_routes$)\;
            \For {each $c_i \in candidate\_routes$}{
                $C \gets C \cup {c_{i}}$\;
            }
        }
    }
    return $C$\;
    \SetKwFunction{Fun}{BidirectionalDijkstra}
    \SetKwProg{Fn}{Function}{:}{}
    \Fn{\Fun($G^v$, $o$, $d$)}{
    
        \While {forward search and reverse search meet on vertex $x$}{
        
            forward search from $o$ on the original graph with a label $d_f$;
            
            reverse search from $d$ on the reversed graph with a label $d_r$;    
        }
        
        return route meet on $x$ of cost $d_f + d_r$\;
    }
    \SetKwFunction{Fun}{Augmentation}
    \SetKwProg{Fn}{Function}{:}{}
    \Fn{\Fun($G^p$, $physical\_routes$)}{
        $new\_routes \gets \emptyset$
        \For {$physical\_route$ $\in$ $physical\_routes$}{
            \For {$route\_segment$ $\in$ $physical\_route$}{
                $new\_route \leftarrow$ replace $route\_segment$ to another edge in $G^p$ with same origin station and destination station\;
                $new\_routes \gets new\_routes \cup \{new\_route\}$
            }
        }
        return $new\_routes$\;
    }
\end{algorithm}

Take \figref{fig:graph_construction} as a running example.
Assume $p_1$ as the origin station and $p_6$ as the destination station, bidirectional Dijkstra's algorithm is first performed on three virtual graphs in parallel. Assume three route candidates on virtual graph $(v^{p_1}_1,v^{p_6}_1))$, $(v^{p_1}_1,v^{p_5}_2,v^{p_6}_2))$ and $(v^{p_1}_1,v^{p_2}_2,v^{p_6}_2)$ are found, three route candidates are translated to $(p_1, p_6)$ in Line 1, $(p_1, p_5, p_6)$ in Line 1 then Line 2, and $(p_1,p_2,p_6)$ in Line 3 then Line 2, via the physical graph. After that, each route segment is replaced by alternative transportation lines to augment route candidate. For example, in the route $(p_1,p_5,p_6)$, the segment $(p_5,p_6)$ in Line 2 can be replaced to Line 1 and will be filtered out later since the new route is identical to $(p_1,p_6)$ in Line 1. As a result, three feasible route candidates are returned.

\section{Route candidate ranking}\label{sec:rank}
In this section, we describe details of the \emph{route candidate ranking} module, including the primary ranking and the re-ranking. The primary ranking derives a diversified route candidate subset whereas the re-ranking orders route based on users' preference under dynamic travel context.

\subsection{Primary ranking}
In general, the route candidate generation produces a route candidate set contains over $50$ route candidates, which is time-consuming to apply a full-fledged machine learning based ranking model directly. Therefore, we first employ a light-weight primary ranking to reduce the size of the route candidate set further.
Concretely, primary ranking reduces the size of the route candidate set in three steps.
First, filter out inferior route candidates, such as similar routes, detour routes, and routes with bad transfer combinations~(\eg metro-bus-metro).
Second, group route candidates based on transport modes, such as bus, metro, mixed transport modes and so on. The grouping step guarantees the diversity of the final route candidate set.
Third, sort routes in each group according to a cost function. The cost function considers multiple factors such as travel time, distance, and the walk distance. The winning route candidates in each group will be passed for re-ranking.
Note that all preserved route candidates will be visualized in front-end, re-ranking will further decide the order of each route candidate.
In \polestar, the primary ranking reduces the route candidate set size to $5$-$7$. 
The primary ranking step can be done in $100$ms.

\subsection{Re-ranking}
After a smaller route candidate set is obtained, we apply a more expensive machine learning based re-ranking model. In this phase, more complex information such as real-time bus, available ticket, and the ticket price are incorporated. We first describe situational features we construct for re-ranking.

\subsubsection{Feature construction}\label{sec:feature}
Feature engineering is a critical step to build an expressive ranking model. Proper feature engineering process can improve the model performance and speed up the convergence of optimization~\cite{zheng2018feature}. Based on explorative data analysis in Section~\ref{sec:data}, we construct five categories of situational features: route features, spatial features, temporal features, meteorological features, and augmented features.

\textbf{Route features}. For each route, we extract \emph{ETA}, \emph{Estimated waiting time}, \emph{Price}, \emph{Ticket availability}, \emph{Road network distance}, \emph{Road congestion index}, \emph{Start walk distance}, \emph{End walk distance}, \emph{On transport distance} and \emph{Number of transfer} from plan records.  
The \emph{Road network distance} is the real travel distance. 
The \emph{Road congestion index} is the congestion score of each route. For transportation not on road network~(\eg metro), the index is set to zero.
The \emph{Start walk distance}~(resp. \emph{End walk distance} ) is the walk distance in the beginning~(resp. in the end).
The \emph{Estimated waiting time} is calculated based on real-time bus information and bus time table.

\textbf{Spatial features}. User's preference at different locations may vary.
We first extract the \emph{city} and the \emph{district} the origin and destination belongs to. Based on the POI data, we extract \emph{Primary POI category} and \emph{Secondary POI category}.
Similar to station binding, we partition the city into a set of grid. For each origin and destination, we further construct statistical features for corresponding grid.
Specifically, we construct the \emph{regional POI distribution} vector, in which each dimension indicate the POI count of each POI category. We further compute  \emph{regional transportation station distribution}, in which each dimension represents the count of transport stations~(\eg bus stations, metro stations, \etc). We also compute \emph{road network density} and \emph{station density} for each grid.
Note that the transportation station data used for re-ranking is same as used for PTG compilation.

\textbf{Temporal features}. The user preference at a different time may also differ. For example, the metro may be a better choice at morning rush hour, whereas the night bus is a possible choice at midnight.
We construct \emph{Hour}, \emph{Minute}, \emph{Day of week}, \emph{Day of month}, \emph{Holiday}, \emph{Route in service} as temporal features. 

\textbf{Meteorological features}. We construct meteorological features from the meteorological dataset. We extract \emph{Weather}, \emph{Temperature}, \emph{Wind speed}, \emph{Wind direction} and \emph{Air Quality Index}~(AQI) as the meteorology features. Specifically, \emph{Weather} and \emph{Wind direction} are categorical features whereas \emph{Temperature}, \emph{AQI}, \emph{Wind speed} and \emph{Humidity} are numerical features.
The weather is categorized as Sunny, Rainy, Overcast, Cloudy, Foggy and Snow. We discretize wind direction to 16 categories. The AQI is an integer value. There are 13 wind strength levels from 0 to 12. The AQI is an integer that represents the air pollution level.

\textbf{Feature augmentation}. We further augment features from two aspects.
First, we compute statistical features to characterize the distribution of each feature.
For route list, we compute $Min$, $Max$ and $Average$ of each route, and compute the difference between basic features and statistical features. 
For example, for distance we compute \emph{Max road network distance}, \emph{Min road network distance}, and \emph{Average road network distance} over all routes in the route list. We compute $ETA-Min ETA$ to measure the relative travel time advantage of a route.
Second, we combine features from different domains to build combinational features. 
For example, we combine route features and temporal features to capture the route statistics at different time period~(\ie hour, day of the week).
We combine route and spatial features to capture the dependency between the POI category and transport mode preference.
Combinational features further capture correlations between each feature category.

\subsubsection{The model}
Given a query $q_i$, a list of route candidates $\mathbf{X}^i=\{\mathbf{x}^i_1, \mathbf{x}^i_2, \dots\}$, the re-ranking model aims to decide the order of each route candidate.
Specifically, each $\mathbf{x}^i_j$ is represented as a $m$ dimensional feature vector, \eg ETA, destination POI type, and weather.
We propose a pair-wise learning to rank model based on GBDT~(Gradient Boosting Decision Tree)~\cite{friedman2001greedy}.
GBDT has proven performs well on tasks with sparse and high-dimensional features. In the learning procedure, the GBDT generates a set of tree classifiers $\mathcal{G}=\{g_1(\cdot),g_2(\cdot),\dots,g_k(\cdot)\}$ sequentially. The rationale behind GBDT is that the model is able to construct a robust classifier by using an ensemble of weak classifiers $g(\cdot)$ to generate the final prediction result
\begin{equation}~\label{equ:gbdtpred}
\hat{y}=f(\mathbf{x}^i)=\sum^k_{j=1}g_j(\mathbf{x}^i),g_j\in \mathcal{F},
\end{equation}
where $\hat{y}$ is the estimated result of instance $\mathbf{x}^i$, $g_j(\cdot)$ is a tree classifier learned in step $j$.

Rather than training each route candidate individually, we build a more informative training set that includes the relative preference relationship.
Specifically, given $q_i$ and $\mathbf{X}^i$, we construct training instances as 
\begin{equation}~\label{equ:rankset}
\mathcal{S}=\{(\mathbf{x}_1^i, \mathbf{x}_2^i)|\mathbf{x}_1^i \succ \mathbf{x}_2^i, i=1,\dots,|\mathbf{X}^i|\},
\end{equation}
where $\succ$ indicate $\mathbf{x}_1^i$ is a more preferable choice than $\mathbf{x}_2^i$ for $q_i$. In fact, $\mathcal{S}$ is a partially ordered set. The partial order relationship is derived by the order of user feedback records. For example, if a user add $\mathbf{x}_1^i$ to favorite before $\mathbf{x}_2^i$ in same query, we have $\mathbf{x}_1^i \succ \mathbf{x}_2^i$.
With the training set, we define the pair-wise ranking loss function as
\begin{equation}~\label{equ:rankloss}
L(q_i, \mathcal{S})=\frac{1}{2}\sum^N_{j=1}(max\{0,\tau - (f(\mathbf{x}_1^i)-f(\mathbf{x}_2^i))\})^2 -\lambda_1\tau^2 + \frac{\lambda_2}{2}\|\mathbf{X}^i\|_2,
\end{equation}
where $f(\cdot)$ is the expected ranking function need to learn. To avoid a trivial optimal $f(\cdot)$ to be obtained, \ie a constant function, we further add a constant gap $\tau\in (0,1]$ to the loss function. $\lambda_1$ and $\lambda_2$ are hyper-parameters for the constant gap and the $L2$ regularization, respectively. We introduce $L2$ regularizer into the loss function to alleviate overfitting in model training.

Given the above loss function, a functional gradient descent~\cite{zheng2007regression} is then applied to optimize the ranking function.
In step $k$, a gradient boosting tree $g_k(\cdot)$ is generated, the ranking function is updated as
\begin{equation}~\label{equ:ranfunc}
f_k(\mathbf{x}^i)=\frac{kf_{k-1}(\mathbf{x}^i)-\beta g_k(\mathbf{x}^i)}{k+1},
\end{equation}
where $\beta$ is the learning rate.

\section{Deployment}\label{sec:deploy}
In this section, we discuss several important implementation and deployment issues,
in both the offline processing phase and the online processing phase.

\subsection{Offline processing}
The offline processing handles four major tasks, the \emph{data management}, the \emph{transportation graph compilation}, the \emph{station cache update}, and the \emph{re-ranking model training}.

\textbf{Data management}.The \emph{Data warehouse} stores all datasets on a Hadoop cluster. The datasets can be categorized into static datasets and dynamic datasets. Static datasets include transportation station data,  transportation line data, road networks, and  POI data, whereas dynamic datasets include query log data, real-time bus data, and traffic condition data. 
Static datasets are updated periodically (from day to months), and dynamic datasets are updated in real-time. 

\textbf{Transportation graph compilation}. Real-world transportation systems are highly dynamic. Each day, new transportation lines open and existing transportation lines are cancelled or adjusted. We propose a graph compilation pipeline to automatic the PTG updating process. In each day, the up-to-date transportation station data, transportation line data, road network data, and a snapshot of traffic condition data are loaded into the data warehouse. The mapping between transportation stations and transportation lines is recompiled to update physical graphs; the mapping between transportation lines and road network is recompiled to update the edge weight of virtual graphs.

\textbf{Station cache update}. Similar to the transportation graph compilation pipeline, the station cache used for station binding is also updated every day to accurately measure the cost to each station. The station cache is stored as a hash table in a Redis database.

\textbf{Re-ranking model training}. 
The re-ranking model relies on multiple large-scale heterogeneous data sets. 
We employ a distributed platform, Bigflow~(http://bigflow.baidu.com), for the data preprocessing.
Specifically, we first integrate each data set into a large fact table by using the $JOIN$ operator, then transform the fact table to a feature table, as described in Section~\ref{sec:feature}. All numerical features are scaled to $[0,1]$ and all categorical features are processed as one-hot encoded vectors.
Once the input feature is ready, we employ XGBoost~\cite{chen2016xgboost}, a high performance distributed gradient boosting library for model training. We use 5-fold cross-validation, and the re-ranking model is updated on a daily basis. The model takes the two most recent months of data as input to exclude seasonal changes. We use a server with 64 Intel Xeon E5-2620v4 CPU, 120GB memory, and 2TB disk for model training.

\subsection{Online processing}
At running time, the response time~(\aka service latency) is crucial to user experience.
We adopt BRPC~(https://github.com/brpc/brpc), an open-sourced scalable web service framework for online service.
In online processing, six components are involved.
First, when a query is submitted, the origin and destination are binded to transportation stations.
Second, the routing executor generates a set of feasible route candidates.
Third, the primary ranking is then applied to select a small set of route candidates.
Fourth, the feature vector for re-ranking is retrieved from the data or collected from other online services.
Fifth, the final order of each route is decided by the re-ranking model.
Finally, each route is returned to the frontend associated with auxiliary data such as road congestion and real-time bus information.
The PTG, the station cache, and the re-ranking model are duplicated in multiple data centers through BRPC. These data centers are distributed in different provinces in China to reduce the latency from different geographical locations and balance the workload.

\section{Experiments}\label{sec:experiments}
In this section, we conduct extensive experiments to evaluate: (1) the efficiency of \polestar, including the running time of route candidate generation and route candidate ranking, (2) the effectiveness of \polestar, including the overall ranking performance and the feature importance analysis, and (3) the online assessment.

\begin{figure}[t]
	\vspace{-1ex}
    \begin{minipage}{1.0\linewidth}
        \centering
        \subfigure[{\small Routing on Shanghai}]{\label{exp:intra-shanghai}
        \includegraphics[width=0.48\textwidth]{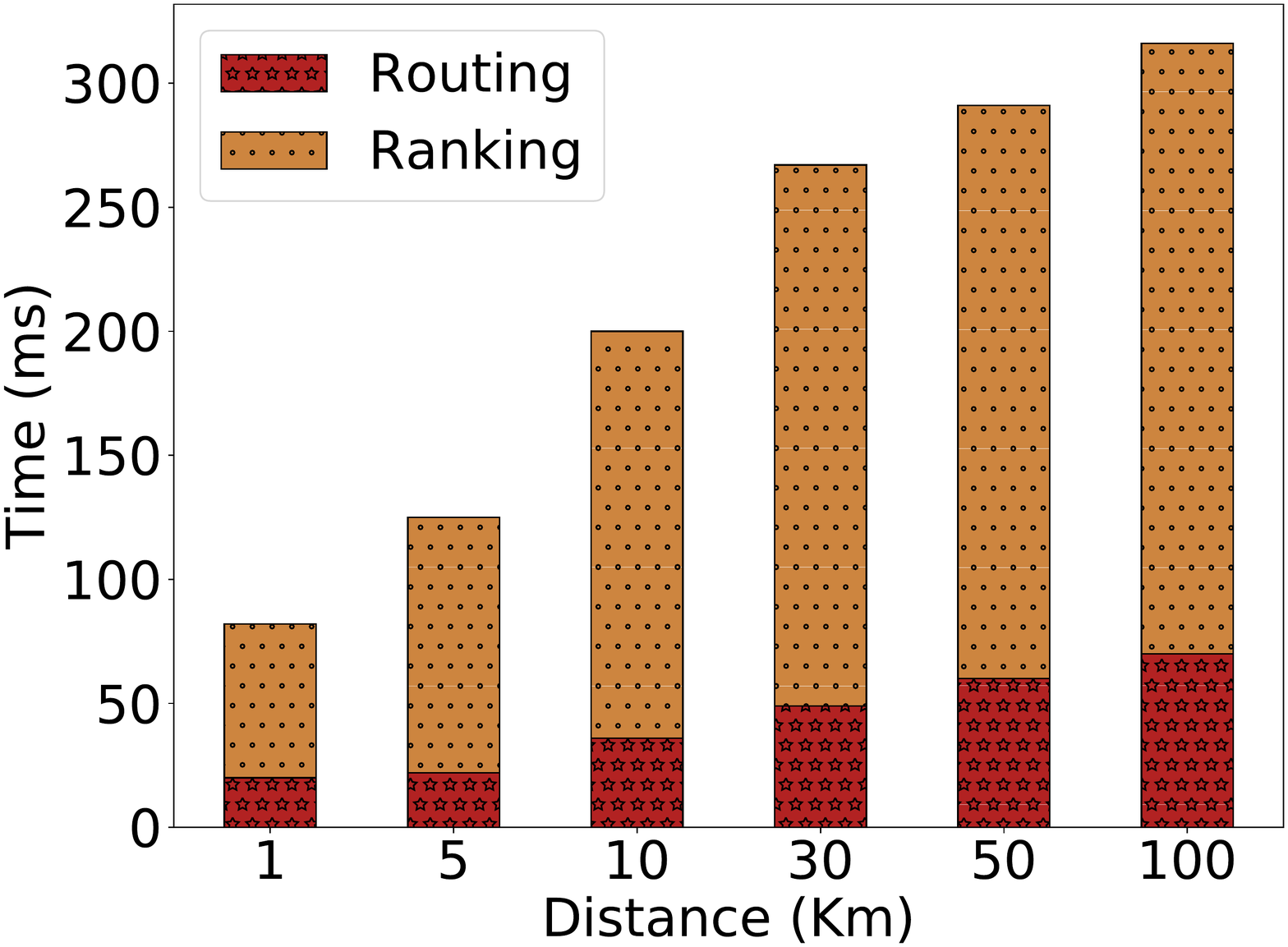}}
        \subfigure[{\small Routing on Guangzhou}]{\label{exp:intra-guangzhou}
        \includegraphics[width=0.48\textwidth]{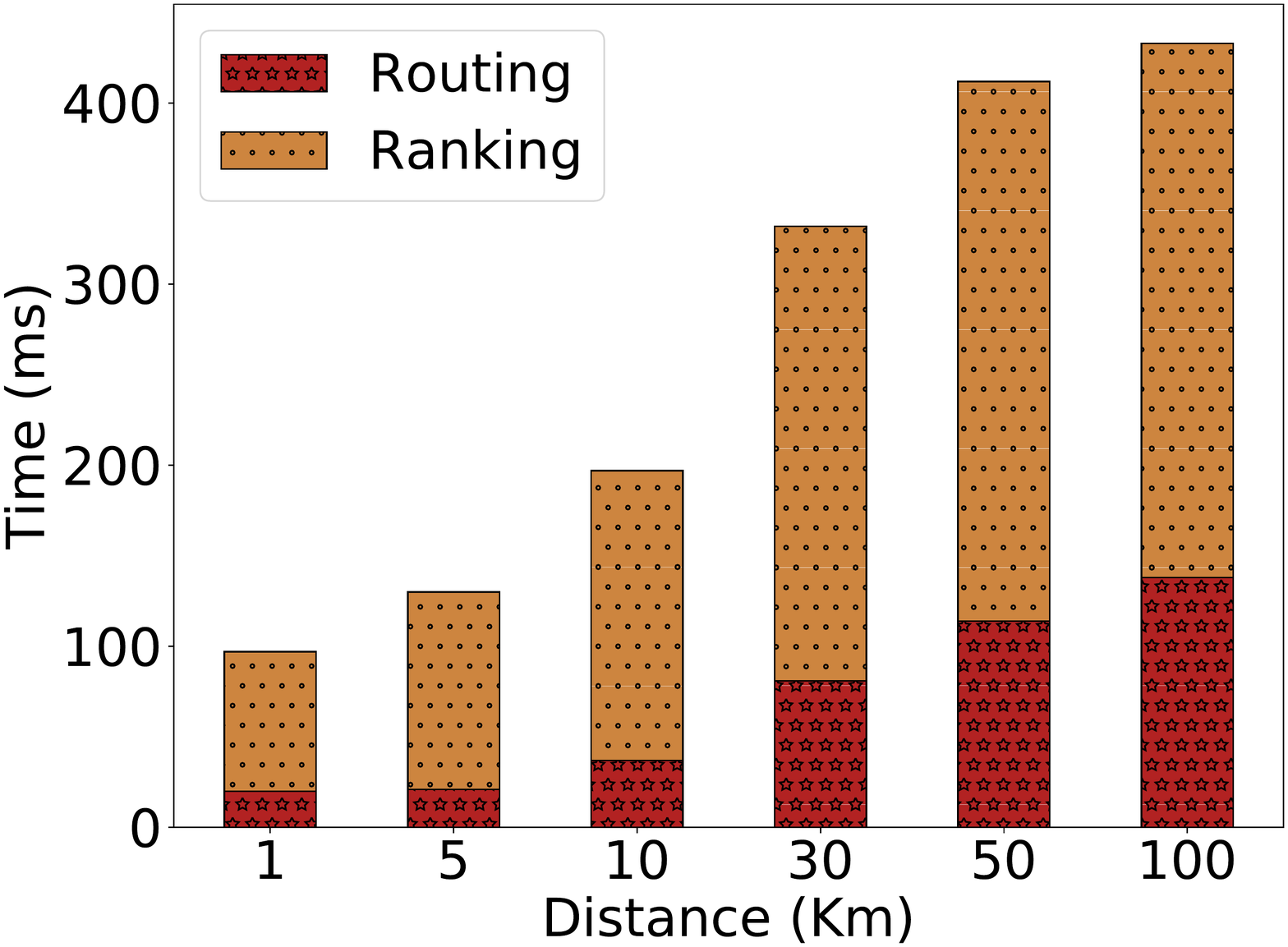}}
        \vspace{-3ex}
    \end{minipage}
    \caption{Running time v.s. distance.}
    \vspace{-12pt}
    \label{exp:running_time}
\end{figure}

\subsection{Efficiency}

We first evaluate the overall running time of \polestar.
Specifically, we exam the performance of \polestar for processing queries of different distances, including both end-to-end and component-wise running time. 
For each distance interval, we randomly extract $1,000$ queries from the log and calculate the averaged running time. 

\figref{exp:intra-shanghai} and \figref{exp:intra-guangzhou} report the running time of queries on Shanghai and Guangzhou. 
We observe that \polestar achieves promising running time for queries of different distances. For queries shorter than $10$~Km, the overall running time is less than $200$~ms. When query distance increases to $100$~km, the running increases to $300$~ms, indicating \polestar achieves sub-linear scalability in terms of query distance.
In \figref{exp:intra-shanghai} and \figref{exp:intra-guangzhou}, we break the running time into routing and ranking phases for further comparison.
It can be seen that the routing phase takes less than $50$ms for all queries, whereas the running time of ranking varies from $50$ms to $200$ms. This indicates that the ranking phase can be further optimized to reduce the overall latency.
Overall, \polestar achieves promising efficiency performance.

\subsection{Effectiveness}
Then we evaluate the overall ranking performance of \polestar and analyze the usefulness of our features.

\textbf{Metric}.
We adopt Normalized Discounted Cumulative Gain (NDCG)~\cite{wang2013theoretical,sigir2020zixuan}, a widely used metric in search engines and recommender systems, to evaluate the ranking performance. 
In this paper, we compare NDCG$@1$, NDCG$@3$ and NDCG$@5$.

\textbf{Baselines}.
We compare \polestar with three statistical baselines and three machine learning based baselines. \textbf{Shortest} rank route candidates according to the overall route distance. \textbf{Fastest} rank route candidates according to the overall travel time. \textbf{Least transfer} rank route candidates according to the number of transfer of each route. \textbf{LR} rank route candidates via the well-known logistic regression model. The input feature is same as \polestar described in Section~\ref{sec:rank}. \textbf{GBDT} uses the same learning model in \polestar, the only difference is GBDT employs a point-wise cross-entropy loss function~\cite{murphy2012machine}. \textbf{DNN} constructs deep neural network contains two fully connected hidden layers as in~\cite{arora2019hard}, applies Relu as the activation function, and employ Adam for optimization.

\subsubsection{Overall ranking result}

\begin{table}[t!]
\centering
\vspace{-3ex}
\begin{small}
\caption{Overall ranking performance.}
\vspace{-8pt}
\begin{tabular}{c | c | c c c  } \hline
& \textbf{Model}
 & NDCG$@1$ & NDCG$@3$  &  NDCG$@5$ \\ \hline \hline
\multirow{7}{*}{\textbf{\shanghai}} & Shortest & 0.428 &0.599 & 0.803 \\
& Fastest & 0.306 & 0.513 & 0.761 \\
& Least transfer & 0.778 & 0.853 & 0.926 \\
& LR & 0.878 & 0.902 & 0.955  \\
& GBDT & 0.921 & 0.922 & 0.966 \\
& DNN & 0.9261 & 0.9189 & 0.9666 \\
& \polestar & \textbf{0.937} & \textbf{0.959} & \textbf{0.979} \\ \hline
\multirow{7}{*}{\textbf{\guangzhou}} & Shortest & 0.472 &0.632 & 0.818 \\
& Fastest & 0.334 & 0.540 & 0.771 \\
& Least transfer & 0.832 & 0.889 & 0.944 \\
& LR & 0.887 & 0.91 & 0.958 \\
& GBDT & 0.916 & 0.92 & 0.965 \\
& DNN & 0.9245 & 0.9205 & 0.9669 \\
& \polestar & \textbf{0.934} & \textbf{0.954} & \textbf{0.977} \\ \hline
\end{tabular}
\label{table-overall}
\end{small}
\vspace{-5pt}
\end{table}

\tabref{table-overall} depicts the overall ranking performance of \polestar and six baselines with respect to NDCG$@k$. 
As can be seen, \polestar significantly outperforms six baselines on both \shanghai and \guangzhou using all three NDCG metrics. 
Specifically, \polestar achieves $0.93+$ NDCG$@1$ score on both datasets, indicating that most routes with user feedbacks are successfully ranked at top-$1$ in our engine.
Besides, we observe that all machine learning based models achieve better ranking performance than statistical based methods, which further proves that our tailored feature construction procedure successfully captures dynamic context factors in user queries.
Moreover, the performance of \polestar is ($1.6\%$, $3.7\%$, $1.3\%$) and ($1.8\%$, $3.4\%$, $1.2\%$) higher than GBDT on two datasets on (NDCG$@1$, NDCG$@3$, NDCG$@5$), demonstrating the power of our pair-wise ranking function.
Finally, we observe \polestar achieves slightly better results than DNN, further illustrate the effectiveness of tree based models on capturing non-linear dependencies. Besides, the computational cost of \polestar is significantly lower than deep neural network, which is an important advantage of \polestar as an online service.

\subsubsection{Feature importance analysis.}

\begin{table}[t]
\centering
\begin{small}
\vspace{-3ex}
\caption{Top-$10$ features ranked by information gain.}
\vspace{-8pt}
\begin{tabular}{c | c | c} \hline
Rank & Feature name & Relative gain\\ \hline \hline 
1 & ETA& 1 \\
2 & ETA - Min ETA & 0.899 \\
3 & Walk distance & 0.785 \\
4 & Walk distance - Min walk distance & 0.654 \\
5 & On transport distance & 0.612 \\
6 & Hour & 0.609 \\
7 & Number of transfer & 0.563 \\
8 & On transport distance & 0.489\\
9 & End walk distance & 0.446 \\
10 &Start walk distance & 0.423\\
\hline
\end{tabular}
\label{table-feature-importance}
\end{small}
\vspace{-5pt}
\end{table}

To evaluate the effectiveness of our features, \tabref{table-feature-importance} reports the top-$10$ most important features in \polestar.
We rank features by information gain~\cite{louppe2013understanding}. Higher information gain means higher frequency the feature used to split nodes in each boosting trees. As we can see, all features are route related features except the $6$th feature. The $6$th feature is a context feature describes the query time. Travel time factors are top-$2$ features, which validates our intuition that time is the most important factors when the user plans a trip.
Moreover, the $2$nd and $4$th features are augmented features, illustrate the effectiveness of our feature augmentation procedure.
Finally, we observe that the relative gain of all top-$10$ features are higher than $0.4$, indicating no feature dominates the re-ranking process.

\subsection{Online assessment}
\begin{table}[h]
\centering
\begin{small}
\vspace{-3ex}
\caption{Online user interview result.}
\vspace{-3ex}
\begin{tabular}{ c | c | c | c | c } \hline
City & G & S & B & Improvement  \\ \hline \hline
\shanghai & 123 & 62 & 15 & $24\%$ \\ \hline
\guangzhou & 52 & 40 & 8 & $32\%$ \\ \hline
\end{tabular}
\label{table-abtest}
\end{small}
\vspace{-3pt}
\end{table}
In the production environment, we observe \polestar achieves $9.4\%$ relative improvement of user click ratio, where the user click ratio is defined as the number of clicks over the number of \polestar routed queries. Besides, \polestar reduces the size of the transportation graph to 3.3 GB, which is only $15.7\%$ as large as the previous one. Moreover, \polestar improves the single server QPS~(query per second) from 30 to 40, achieving a $33.3\%$ improvement. 

To evaluate the quality of new ranking results, we conduct user interview one week after \polestar fully deployed. 
Specifically, we published survey questionnaires in Baidu Maps.
In the questionnaire, we set three level ranking category, $G$, $S$, $B$, where $G$ stands for better than the previous ranking result, $S$ stands for same as the previous ranking result, and $B$ stands for worse than the previous ranking result. 
The user interview result in \shanghai and \guangzhou is reported in Table~\ref{table-abtest}. 
The improvement of the new ranking result is defined as 
$\frac{\#~of~G - \#~of~B}{\#~of~feedbacks}$.
Overall, \polestar achieves $24\%$ and $32\%$ gain in \shanghai and \guangzhou, respectively.

\section{Related work}\label{sec:related}
Transportation routing can be partitioned into \emph{model-free} routing and \emph{model-based} routing.

\textbf{Model-free routing} aims to build efficient algorithms to answer specific queries, \eg earliest arrival, latest departure, and shortest duration~\cite{wang2015efficient}. For public transportation, the network is usually formalized as a timetable graph~\cite{efentakis2016scalable}.
On the one hand, efficient index structures such as hub labeling~\cite{abraham2012hierarchical} and contraction hierarchy~\cite{geisberger2008contraction} are proposed to speed up such shortest path search. 
On the other hand, Funke \etal~\cite{funke2015personalized} and Sacharidis~\cite{Sacharidis:2017:FMP:3139958.3140029} study personalized shortest path routing that supports various optimization criteria. 
Delling \etal~\cite{delling2012computing,dibbelt2016engineering} study the routing problem with multiple transport mode choices.
Efficiency and reachability are major concerns of above model-free routing algorithms in city-wide, whereas the user preference and the dynamic travel context are overlooked.

\textbf{Model-based routing} employs learning or mining models to improve routing performance. For example, MPR~\cite{chen2011discovering} discovers sequential patterns from the user trip history and proposes efficient indices to speed up the retrieval of such patterns. 
T-drive~\cite{yuan2010t} improves route quality based on the taxi driver's intelligence and Dai~\cite{dai2015personalized} recommend routes from historical trajectories rather than ad-hoc shortest path search.
Moreover, Wang \etal~\cite{wang2019empowering} incorporates neural network into $A^*$ algorithm for personalized route recommendation. 
Hydra~\cite{Hydra} introduces a data-driven approach that incorporates a network embedding model~\cite{hao2019trans2vec} for multi-modal route recommendation. 
All of the above approaches focus on finding better driving or traveling routes in a single city and are not designated for public transportation routing.

\section{Conclusion}\label{sec:conclusion}

In this paper, we presented \polestar, an intelligent, efficient and national wide public transportation engine.
We first proposed PTG, a unified graph structure to integrate heterogeneous public transportation lines.
Based on PTG, a station binding method and bidirectional Dijkstra algorithm are introduced for efficient route candidate generation.
A two-pass ranking framework is then proposed for route candidate selection and ranking. The re-ranking model constructs a rich set of situational features and adopts a learning to rank model to capture user preference under dynamic travel context.
\polestar has been deployed and tested at scale on Baidu Maps and answers a hundred millions of queries each day.
We believe the proposed framework is not limited to public transportation routing, but also can be referenced for developing other large-scale transport routing engines.
We share our practical experience on how to build such a production level routing engine and hope to provide useful insights to communities in both academia and industry.

\small
\bibliographystyle{ACM-Reference-Format}
\bibliography{ref}

\end{document}